\documentclass{article}

\usepackage{microtype}
\usepackage{graphicx}
\usepackage{subfigure}
\usepackage{booktabs} 
\usepackage{siunitx}
\usepackage{hyperref}
\usepackage{amsmath}
\usepackage{amssymb}
\usepackage{mathtools}
\usepackage{amsthm}

\usepackage[accepted]{icml2025}

\usepackage[capitalize,noabbrev]{cleveref}

\theoremstyle{plain}

\theoremstyle{definition}

\theoremstyle{remark}

\DeclareSIUnit\angstrom{\text {Å}}
\usepackage[textsize=tiny]{todonotes}

 \icmltitlerunning{Beyond Atomic Geometry Representations in Materials Science}

\begin{document}

\twocolumn[

\icmltitle{Beyond Atomic Geometry Representations in Materials Science:\\ A Human-in-the-Loop Multimodal Framework}




\begin{icmlauthorlist}
\icmlauthor{Can Polat}{yyy}
\icmlauthor{Erchin Serpedin}{yyy}
\icmlauthor{Mustafa Kurban}{tamuq,ankara}
\icmlauthor{Hasan Kurban}{comp}
\end{icmlauthorlist}

\icmlaffiliation{yyy}{Dept. of Electrical and Computer Engineering, Texas A\&M University, College Station, TX 77843, USA}
\icmlaffiliation{comp}{College of Science and Engineering, Hamad Bin Khalifa University, Doha, Qatar}
\icmlaffiliation{tamuq}{Dept. of Electrical \& Computer Engineering, Texas A\&M University at Qatar, Doha, Qatar}
\icmlaffiliation{ankara}{Dept. of Prosthetics and Orthotics, Ankara University, Ankara, Turkey}

\icmlcorrespondingauthor{Hasan Kurban}{hkurban@hbku.edu.qa}
\icmlcorrespondingauthor{Mustafa Kurban}{kurbanm@ankara.edu.tr}

\icmlkeywords{Benchmark, Crystals, Multimodal, DFT, LLM, Machine Learning, ICML}

\vskip 0.3in
]



\printAffiliationsAndNotice{}  

\begin{abstract}
Most materials science datasets are limited to atomic geometries (e.g., XYZ files), restricting their utility for multimodal learning and comprehensive data-centric analysis. These constraints have historically impeded the adoption of advanced machine learning techniques in the field. This work introduces \textbf{MultiCrystalSpectrumSet} (MCS-Set), a curated framework that expands materials datasets by integrating atomic structures with 2D projections and structured textual annotations, including lattice parameters and coordination metrics. MCS-Set enables two key tasks: (1) multimodal property and summary prediction, and (2) constrained crystal generation with partial cluster supervision. Leveraging a human-in-the-loop pipeline, MCS-Set combines domain expertise with standardized descriptors for high-quality annotation. Evaluations using state-of-the-art language and vision-language models reveal substantial modality-specific performance gaps and highlight the importance of annotation quality for generalization. MCS-Set offers a foundation for benchmarking multimodal models, advancing annotation practices, and promoting accessible, versatile materials science datasets. The dataset and implementations are available at \url{https://github.com/KurbanIntelligenceLab/MultiCrystalSpectrumSet}.
\end{abstract}

\section{Introduction}
Materials science drives innovation in energy storage, catalysis, and microelectronics, yet progress remains limited by the challenge of mapping atomic geometry to material function~\cite{shen2022machine,choudhary2022recent,huang2024crystal,jaafreh2022crystal}. While first-principles methods such as density functional theory (DFT) offer accurate predictions of ground-state properties, their performance depends heavily on high-quality initial structures and exhaustive sampling of the configurational landscape—constraints that scale poorly with system complexity~\cite{orio2009density,cohen2012challenges}. Semi-empirical surrogates like density functional tight binding (DFTB) reduce computational cost but still require extensive geometry optimization, sustaining a “structure bottleneck” that hinders high-throughput discovery \cite{hourahine2007self}.

Recent advances in machine learning have introduced data-driven surrogates, including graph neural networks (GNNs), diffusion models, and large language models (LLMs), to predict energies or synthesize plausible crystal configurations~\cite{xie2021crystal,jiao2023crystal,antunes2024crystal,li2024machine,hessmann2025accelerating}. These methods have been enabled by curated datasets such as \textsc{Perov-5}~\cite{castelli2012computational}, \textsc{Carbon-24}~\cite{pickard2006high,pickard2011ab}, and \textsc{MP-20}~\cite{jain2013commentary}. However, two systemic issues persist. First, heterogeneous data curation introduces sampling bias and spurious correlations that limit model robustness and out-of-distribution performance~\cite{davariashtiyani2024impact}. Second, most existing benchmarks are restricted to atomic coordinate data, omitting the multimodal context—such as visual projections and textual descriptors—that human crystallographers routinely use to reason about structure.

\textbf{MultiCrystalSpectrumSet} (MCS-Set) is a human-in-the-loop, multimodal benchmark for evaluating structure–property relationships in materials science. It consists of atomic clusters of Ag, Au, PbS, and ZnO, spanning R6–R10 (0.6 nm to 1 nm) geometries and ranging from 55 to 351 atoms. Each cluster is augmented with 780 unique 3D rotations, producing over 15,600 triplets that align XYZ coordinate files with high-resolution (512\(\times\)512) orthographic projections and structured textual annotations. These annotations include lattice parameters, unit cell volume, density, and nearest-neighbor statistics. Annotation quality is maintained through an expert-in-the-loop process that combines automated structural descriptors with manual review.

MCS-Set supports two core research directions: (1) \textit{multimodal property and summary prediction}, leveraging integrated structural and visual inputs, and (2) \textit{constrained crystal generation} under partial cluster supervision, such as extrapolating R9 structures from R6, R7, R8, and R10 data. The benchmark enables cross-modal evaluation using both geometry-aware and language-based metrics, and facilitates analysis of annotation fidelity, rotational augmentation effectiveness, and modality-specific generalization. By unifying structured, visual, and textual modalities, MCS-Set promotes data-centric practices in materials science and contributes to reproducible, generalizable benchmarks for scientific machine learning.

The rest of the paper is organized as follows. Section \ref{sec:bg} presents background and related work. Section \ref{sec:data} describes the data generation process. Section \ref{sec:tasks} outlines the tasks, implementation, and experiments. Section \ref{sec:limitations} discusses limitations, and Section \ref{sec:conclusion} marks the conclusion.
\begin{figure*}[!t]
    \centering
    \includegraphics[width=0.95\linewidth]{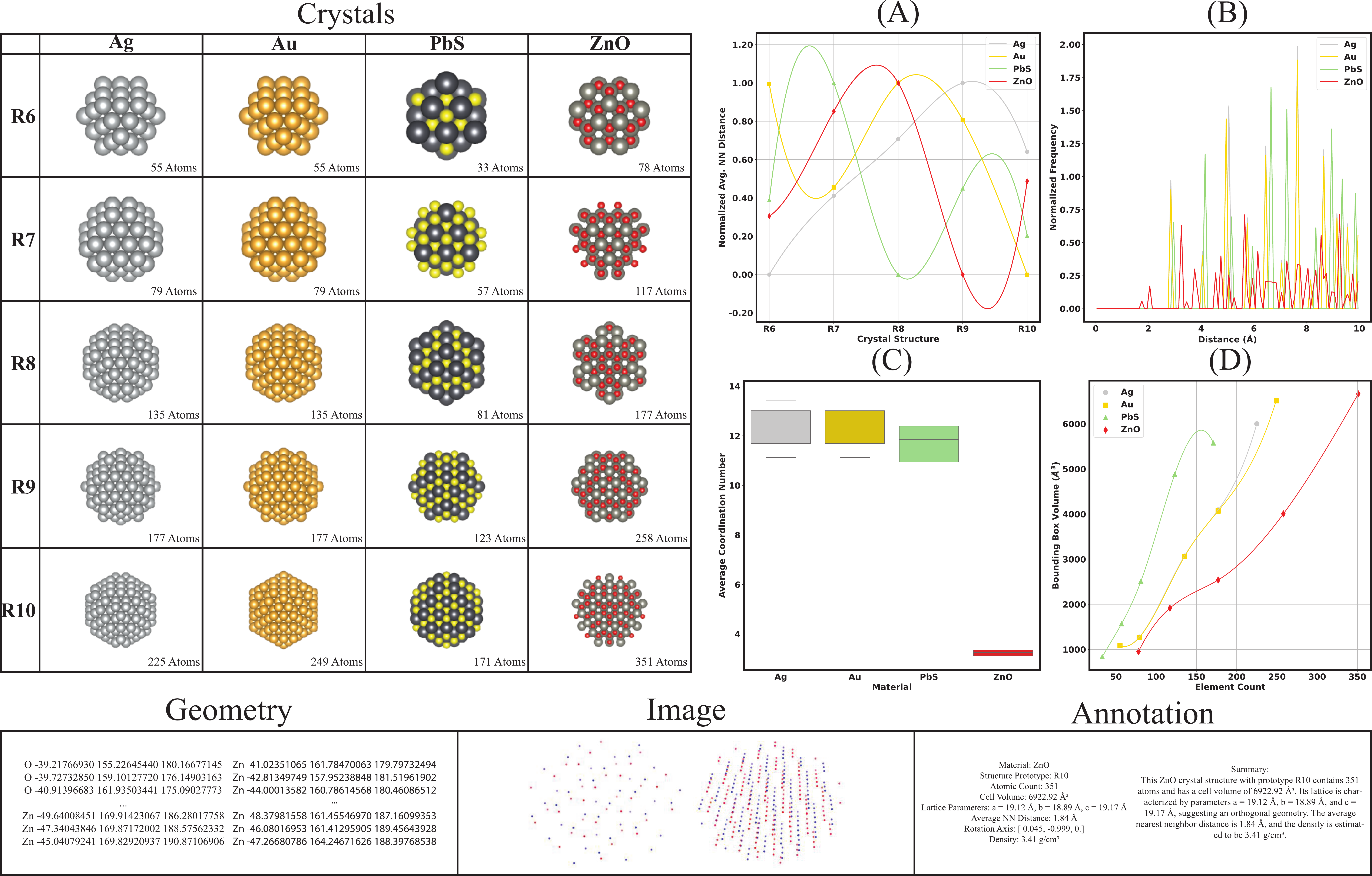}
    \caption{
    Overview of the \textit{MultiCrystalSpectrumSet} (MCS-Set) dataset. \textbf{Top Left:} Atomic clusters of Au, Ag, PbS, and ZnO spanning R6–R10 geometries. \textbf{Top Right:} Structural property distributions across materials. (\textbf{A}) Normalized average nearest-neighbor distances. (\textbf{B}) Radial distribution functions highlighting pairwise atomic distances. (\textbf{C}) Coordination number distributions. (\textbf{D}) Atomic count versus bounding box volume. \textbf{Bottom:} Representative triplets from each modality—XYZ atomic coordinates, high-resolution 2D projections, and structured textual annotations—demonstrating the dataset's multimodal alignment. The figure illustrates both the diversity of crystal structures and the design of MCS-Set to support multimodal learning and benchmarking.
    }
    \label{fig:data_fig}
\end{figure*}

\section{Background}
\label{sec:bg}

This section reviews computational techniques and data resources relevant to crystal structure modeling. Traditional approaches rely on DFT and semi-empirical search methods, while recent generative models learn structural distributions directly from data. Machine learning methods—ranging from convolutional networks to transformers and graph neural networks—encode varying degrees of structural inductive bias suited to materials domains.

Public benchmarks have accelerated progress in property prediction and structure generation, yet most remain limited to raw atomic coordinate tables, lacking aligned multimodal data and transparent curation. This constraint hinders comprehensive data-centric evaluation and the development of generalizable multimodal models. These limitations highlight the need for datasets that integrate multiple modalities and embed human expertise into the annotation process—motivating the design and expert-guided construction of MCS-Set.

\subsection{Crystal-Structure Generation Approaches}

Material properties—mechanical, optical, and electronic—are inherently determined by crystal geometry \cite{bhadeshia2001geometry}. Ab-inito methods such as DFT remain the gold standard for predicting ground-state structures, but their effectiveness depends on high-quality initial coordinates and exhaustive sampling of configurational space. Semi-empirical alternatives, including DFTB, offer improved efficiency and have been extended to broader chemical regimes, including {\it 5f} elements.

To overcome scalability bottlenecks, global optimization strategies have been widely adopted, including simulated annealing~\cite{bertsimas1993simulated}, evolutionary algorithms~\cite{bartz2014evolutionary}, basin hopping, and data-mining heuristics~\cite{huber2020machine}. These methods provide diverse candidate structures but require careful parameterization and substantial computational effort.

\subsection{Machine Learning in Materials Science}

Machine learning has become integral to modern materials discovery pipelines. Convolutional networks~\cite{mishra2023detailed,zheng2018machine}, transformers~\cite{rane2023transformers,liao2023equiformerv2}, GNNs~\cite{schnet,dimenet,du2024densegnn,faenet}, and multimodal encoders~\cite{polat2024multimodal} have been applied to predict formation energies, band gaps, and elastic tensors with near-DFT accuracy.

Recent advances in generative modeling are revolutionizing crystal structure generation by introducing data-driven surrogates. Models like CDVAE~\cite{xie2021crystal}, DiffCSP~\cite{jiao2023crystal}, and EH-Diff~\cite{liu2025equivariant} (an equivariant hypergraph diffusion framework) can explore configurational space significantly faster than traditional search methods. Despite this speed, these approaches are very sensitive to dataset quality, symmetry preservation, and modality alignment. This highlights the need for carefully curated, structurally consistent training data to achieve generalizable crystal generation. Beyond single modalities, generative frameworks are evolving toward multimodal settings, as seen with CrystaLLM~\cite{antunes2024crystal}, which combines structural graphs and text for text-conditioned crystal structure generation. When scaled, graph-based models have been effectively used in high-throughput discovery pipelines, such as DeepMind’s method for identifying stable compounds across vast chemical spaces~\cite{merchant2023scaling}.

Benchmark datasets have been instrumental in driving algorithmic advances across machine learning. Multimodal evaluation suites such as ScienceQA~\cite{lu2022learn}, SciBench~\cite{wang2023scibench}, LabBench~\cite{laurent2024lab}, MMMU~\cite{yue2024mmmu}, TDCM25~\cite{polat2025tdcm25}, and OlympiadBench~\cite{he2024olympiadbench} assess scientific reasoning across images, equations, and text. Complementary domain-specific benchmarks—including MoleculeNet~\cite{wu2018moleculenet}, ChemLit-QA~\cite{wellawatte2024chemlit}, and HoneyComb~\cite{zhang2024honeycomb}—extend multimodal evaluation to chemistry and materials science.

Crystallographic datasets, however, remain narrowly scoped. The Cambridge structural database~\cite{camcd} offers high-quality experimental structures, while CSPBench provides 180 optimized test cells for structure prediction evaluation. CHILI~\cite{friis2024chili} introduces multi-scale inorganic graphs for generative modeling. Despite these advances, most datasets remain limited to atomic coordinates, lack explicit quality audits, and rarely include aligned visual or textual modalities. These constraints hinder multimodal evaluation and restrict systematic study of data-centric interventions—gaps that MCS-Set is designed to address.

\section{Dataset}
\label{sec:data}
The MCS-Set dataset integrates ab initio consistent crystal clusters, rotationally augmented views, pixel-space renderings, and structured geometric descriptors to enable structure-aware, multimodal learning. The subsections below detail the crystal generation protocol, the Fibonacci-sphere sampling strategy for orientation diversity, the 2D projection pipeline, and the construction of structured textual annotations.

\subsection{Crystal Simulations}

Let the three lattice vectors of the experimental unit cell be  
\[
\mathbf{A} \;=\; \bigl[\mathbf{a}_1\;\mathbf{a}_2\;\mathbf{a}_3\bigr]\;\in\;\mathbb{R}^{3\times3}.
\]
A supercell is generated by drawing an integer multiplicity matrix  
\(\mathbf{S}\in\mathbb{Z}_{>0}^{3\times3}\) such that  
\(\det\mathbf{S}\le8\).  Each atomic motif index \(\mathbf{n}\in\mathbb{Z}^3\)  
is mapped to Cartesian coordinates via
\[
\mathbf{r}' \;=\; \mathbf{A}\,\mathbf{S}\,\mathbf{n}.
\]

Silver, gold, and lead-sulfide adopt face-centred-cubic (FCC) symmetry with lattice parameters  
\[
a_{\mathrm{Ag}}=\SI{4.0857}{\angstrom},\quad
a_{\mathrm{Au}}=\SI{4.0780}{\angstrom},\quad
a_{\mathrm{PbS}}=\SI{5.9362}{\angstrom}.
\]
Zinc-oxide crystallizes in the wurtzite structure, parameterized by  
\[
a_{\mathrm{ZnO}}=\SI{3.2495}{\angstrom},\quad
c_{\mathrm{ZnO}}=\SI{5.2069}{\angstrom}.
\]

Near-spherical clusters of radius  
\[
R_k \;=\; 0.2\,k\,a_{\mathrm{Ag}},\qquad k=6,7,8,9,10,
\]
are carved from each FCC supercell.  These clusters contain  
\(N_k\sim k^3\) atoms (ranging from 55 to 351) and are labeled  
\(\mathrm{R}6\) through \(\mathrm{R}10\).  Only atoms satisfying
\[
\bigl\lVert \mathbf{r}' - \mathbf{r}_0 \bigr\rVert_2 \;\le\; R_k,
\]
where \(\mathbf{r}_0\) denotes the center of mass, are retained.  

\subsection{Symmetry Augmentation via Fibonacci Sphere}

To guarantee quasi-uniform coverage of the rotation group $\mathrm{SO}(3)$, each baseline structure is augmented with $N = 780$ orientations. The $i$-th rotation axis $\mathbf{n}_i\in S^2$ is drawn from the Fibonacci lattice \cite{Stanley01101975}:

\begin{align}
\label{eq:fib-axis}
\mathbf{n}_i 
&= \bigl(\sqrt{1 - y_i^2}\,\cos\varphi_i,\; 
        y_i,\; 
        \sqrt{1 - y_i^2}\,\sin\varphi_i\bigr), \\[6pt]
y_i 
&= 1 - \frac{2\,(i + 0.5)}{N}, \\[3pt]
\varphi_i 
&= 2\pi\,i\,\phi,
\end{align}

where $\phi = (\sqrt{5}-1)/2$ is the golden-ratio conjugate. A fixed angle $\theta = \pi/5$ is then applied via Rodrigues’ formula \cite{bezerra2021euler}:

\begin{align}
\label{eq:rodrigues}
\mathbf{R}_i(\theta)
&= \mathbf{I}_3
  + \sin\theta\,\bigl[\mathbf{n}_i\bigr]_{\times}
  + (1 - \cos\theta)\,\bigl[\mathbf{n}_i\bigr]_{\times}^{2}, \\[6pt]
\label{eq:skew}
\bigl[\mathbf{n}\bigr]_{\times}
&=
\begin{pmatrix}
   0    & -n_{z} &  n_{y} \\
  n_{z} &   0    & -n_{x} \\
 -n_{y} &  n_{x}  &   0
\end{pmatrix}.
\end{align}

Coordinates are then updated as
\begin{equation}
\mathbf{r}'' = \mathbf{R}_i(\theta)\,\mathbf{r}',
\end{equation}

and the worst-case angular discrepancy between neighbouring axes scales as $\mathcal{O}(N^{-1})$, ensuring dense sampling without redundancy.

\subsection{Two-Dimensional Representations}

An oriented structure is converted to pixel space by homogeneous coordinates \(\hat{\mathbf{r}}=(\mathbf{r}^{\prime\prime},1)^{\!\top}\). For orthographic projection,
\[
\mathbf{P}_{\text{ortho}}=
\begin{pmatrix}
1 & 0 & 0 & 0\\
0 & 1 & 0 & 0\\
0 & 0 & 0 & 1
\end{pmatrix},
\quad
\hat{\mathbf{u}}=\mathbf{P}_{\text{ortho}}\hat{\mathbf{r}},
\]
yielding constant scale and eliminating perspective distortion. A perspective variant uses the pinhole model
\(
\mathbf{P}_{\text{persp}}=
\begin{psmallmatrix}
f & 0 & 0 & 0\\
0 & f & 0 & 0\\
0 & 0 & 1 & 0
\end{psmallmatrix}
\)
so that \((u,v)=(fx/z,fy/z)\). Rendered images are \(512\times512\) px with atom sizes proportional to covalent radii and colours mapped to element types.   

\subsection{Structured Annotations}

Let \(\mathcal{S}=\{\mathbf{r}_{i}\}_{i=1}^{N}\subset\mathbb{R}^{3}\) be the
atomic set of an oriented cluster.  Four labels are recorded:

\begin{enumerate}\setlength\itemsep{4pt}
\item \textbf{Axis-aligned cell metrics}.  
      \(a=\max r_{x}-\min r_{x},\;
        b=\max r_{y}-\min r_{y},\;
        c=\max r_{z}-\min r_{z}\).

\item \textbf{Cell volume}. \(V=abc\).

\item \textbf{Mean first-neighbour distance}.  
      \(d_{i}=\min_{j\neq i}\lVert\mathbf{r}_{i}-\mathbf{r}_{j}\rVert_{2}\),
      \(\bar{d}=\tfrac{1}{N}\sum_{i}d_{i}\).\,%

\item \textbf{Mass density}.  
      \(M=\sum_{i}m_{i},\;
        \rho=M/V\).

\end{enumerate}

Optionally, the radial distribution function \(g(r)=\frac{1}{4\pi r^{2}\rho}\sum_{i\neq j}\delta(r-\lVert \mathbf{r}_{i}-\mathbf{r}_{j}\rVert_{2})\) is tabulated on a discrete grid to support symmetry-aware contrastive losses. These mathematically explicit descriptors enable precise, differentiable evaluation for both discriminative and generative benchmarks.

\section{Tasks, Implementation, and Experiments}
\label{sec:tasks}

This section formalises the two benchmark tasks released with
MCS-Set, outlines the baseline implementations, and analyses empirical findings.

\subsection{Task 1: Structural-Property Prediction and Summary Generation}
\label{subsec:task1}

\textbf{Objective.} Given one orthographic image
$\mathbf{I}\!\in\!\mathbb{R}^{512\times512\times3}$ and its aligned
XYZ coordinate set
$\mathcal{X}=\{\mathbf{r}_i\}_{i=1}^{N}\subset\mathbb{R}^3$,
the model must regress six scalar properties—
lattice parameters $(a,b,c)$, pseudo-cell volume $V=abc$,
average nearest-neighbour distance $\bar d$,
correlation number,\footnote{Computed as the average coordination
number of each atom.}
and mass density $\rho$—and produce a \emph{concise}
($\le40$-token) natural-language summary.

\textbf{Motivation.} The task evaluates whether \emph{multimodal curation}
(images \emph{and} coordinates) improves lattice-scale reasoning while
simultaneously testing if free-form text faithfully reflects numeric
predictions—an explicit data-quality concern.

\textbf{Metrics.} Let the test set be
\(\mathsf{D}_{\mathrm{test}}
   =\{(\mathbf{x}_i,\mathbf{y}_i,s_i)\}_{i=1}^{|\mathsf{D}|}\),
with targets
\(\mathbf{y}_i=(a,b,c,V,\bar d,\rho)_i^{\!\top}\)
and reference summary \(s_i\).

\textbf{Scalar regression.}
For each of the six components
\(k\in\{1,\dots,6\}\)
\[
\mathrm{MAE}_k
=\frac{1}{|\mathsf{D}|}
  \sum_{i=1}^{|\mathsf{D}|}
  \bigl|\,\hat y_{ik}-y_{ik}\bigr| \; .
\]

\textbf{Surface fluency.}
BLEU-4 is computed with brevity penalty
\(\mathrm{BP}=e^{\max(0,1-\tfrac{r}{c})}\)
and modified \(n\)-gram precisions \(p_n\):
\[
\mathrm{BLEU}_4
=\mathrm{BP}\;\exp\!\Bigl(\tfrac14\sum_{n=1}^{4}\ln p_n\Bigr).
\]
ROUGE-L follows the longest-common-subsequence
\(F_\beta\!\)-measure of Lin.

\textbf{Numeric fidelity (FactScore).} Let \(\mathrm{num}(\cdot)\) return the multiset of decimal numbers
rounded to \(10^{-3}\).
Exact numeric agreement is then
\[
\mathrm{FactScore}
=\frac{1}{|\mathsf{D}|}
  \sum_{i=1}^{|\mathsf{D}|}
  \mathbb{I}\bigl[
    \mathrm{num}(\hat s_i)=\mathrm{num}(s_i)
  \bigr].
\]

\subsection{Task 2 : Crystal Generation from Unseen R-Combinations}
\label{subsec:task2}

\textbf{Objective.} A model observes clusters at radii
$\mathrm{R}6,\mathrm{R}7,\mathrm{R}8$, and $\mathrm{R}{10}$ for a given
chemistry and must synthesise plausible $\mathrm{R}9$ structures that
never appear during training.

\textbf{Motivation.} The held-out radius constitutes a controlled distribution shift,
allowing robustness claims to be benchmarked in a \emph{data-centric}
manner.

\textbf{Generation Metrics.} Assume a test corpus
\(\mathsf D_{\mathrm{test}}
   =\{(\widehat{\mathcal X}_i,
        \mathcal X^{\mathrm{gt}}_i)\}_{i=1}^{|\mathsf D|}\),
where
\(\widehat{\mathcal X}_i
   =\{\hat{\mathbf r}_{ij}\}_{j=1}^{\widehat N_i}\)
is the predicted cluster and
\(\mathcal X^{\mathrm{gt}}_i
   =\{\mathbf r_{ij}\}_{j=1}^{N_i^{\mathrm{gt}}}\)
is the ground truth.

\textbf{Validity} is the fraction of predictions whose minimum
inter-atomic separation exceeds \(0.5\text{\,\AA}\):
\[
\mathrm{Validity}
=\frac1{|\mathsf D|}
 \sum_{i=1}^{|\mathsf D|}
 \mathbb I\!\bigl[
   \min_{j\neq k}
   \Vert\hat{\mathbf r}_{ij}-\hat{\mathbf r}_{ik}\Vert_2
   >0.5\text{\,\AA}
 \bigr]
\times100\%.
\]

\textbf{Atom-count error (ACE)} measures the relative difference in
cardinality:
\[
\mathrm{ACE}
=\frac1{|\mathsf D|}
 \sum_{i=1}^{|\mathsf D|}
 \frac{\bigl|\widehat N_i-N_i^{\mathrm{gt}}\bigr|}
      {N_i^{\mathrm{gt}}}
\times100\%.
\]

When \(\widehat N_i = N_i^{\mathrm{gt}}\) (otherwise the metric is
undefined and reported as \textit{N/A}), two topology-aware scores are
computed.  
The \textbf{root-mean-square deviation (RMSD)} uses Kabsch alignment
\(\mathbf R_i^\star\in\mathrm{SO}(3)\):
\[
\mathrm{RMSD}
=\frac{
   \displaystyle
   \sum_{i:\,\widehat N_i=N_i^{\mathrm{gt}}}
   \Bigl(
     \frac1{N_i^{\mathrm{gt}}}
     \sum_{j=1}^{N_i^{\mathrm{gt}}}
     \bigl\|
       \mathbf R_i^\star\hat{\mathbf r}_{ij}
       -\mathbf r_{ij}
     \bigr\|_2^{2}
   \Bigr)^{1/2}}
  {\#\{i:\widehat N_i=N_i^{\mathrm{gt}}\}}.
\]

The \textbf{match rate (MR)} counts how often the bidirectional Chamfer
distance does not exceed a tolerance \(\varepsilon=0.25\text{\,\AA}\):
\[
\mathrm{MR}
=\frac{
   \displaystyle
   \sum_{i:\,\widehat N_i=N_i^{\mathrm{gt}}}
   \mathbb I\!\bigl[
     \mathrm{Chamfer}(
       \widehat{\mathcal X}_i,\,
       \mathcal X^{\mathrm{gt}}_i)
     \le\varepsilon
   \bigr]}
  {\#\{i:\widehat N_i=N_i^{\mathrm{gt}}\}}
\times100\%.
\]

\subsection{Results}
\label{subsec:results}

\begin{table*}
\centering
\caption{Evaluation metrics for various LLMs on both textual generation and structural property prediction in Task 1. Averaged for 10 different samples from each material and each R-configuration. Best values are in \textbf{bold}, second-best are \underline{underlined}.}
\resizebox{\textwidth}{!}{%
\begin{tabular}{lccccccccccccc}
\toprule
Model & BLEU & ROUGE-1 & ROUGE-2 & ROUGE-L & $\%\Delta$Atoms & $\%\Delta V$ & $\%\Delta a$ & $\%\Delta b$ & $\%\Delta c$ & $\%\Delta \mathrm{NN}$ & $\%\Delta \rho$ & Mat.\ Match & Struct.\ Match \\
\midrule
Claude 3.5 Haiku & \textbf{0.61} & 0.73 & 0.56 & 0.71 & 14.93 & 270.6 & 400.8 & 391.9 & 382.8 & 8.75 & \textbf{40.17} & \textbf{1.00} & \textbf{1.00} \\
Claude 3.5 Sonnet & \underline{0.60} & \textbf{0.75} & \textbf{0.58} & \textbf{0.74} & \textbf{9.370} & \textbf{44.31} & \textbf{8.180} & \textbf{7.510} & \textbf{23.33} & \textbf{1.97} & 47.46 & 0.69 & \textbf{1.00} \\
Deepseek Chat & \textbf{0.61} & \underline{0.74} & \underline{0.57} & \underline{0.73} & 16.75 & 94.72 & 57.60 & 44.49 & 45.38 & 5.63 & 46.43 & \textbf{1.00} & \textbf{1.00} \\
Gemini 2.0 Flash & \textbf{0.61} & \textbf{0.75} & \underline{0.57} & \underline{0.73} & \underline{11.70} & \underline{89.48} & 73.76 & 78.43 & 100.7 & 2.73 & 49.78 & \textbf{1.00} & 0.90 \\
Gemma 3 27b & \textbf{0.61} & \underline{0.74} & \underline{0.57} & 0.72 & 37.34 & 171.6 & \underline{29.89} & \underline{33.92} & 46.85 & 22.6 & 46.65 & \underline{0.99} & \underline{0.99} \\
GPT-4o & 0.40 & 0.51 & 0.39 & 0.50 & 15.93 & 169.2 & 38.46 & 38.19 & \underline{41.97} & 2.78 & \underline{45.96} & 0.67 & 0.67 \\
Grok 2 & \textbf{0.61} & \underline{0.74} & \textbf{0.58} & \underline{0.73} & 20.65 & 1537 & 156.3 & 147.9 & 151.3 & \underline{2.10} & 46.37 & \textbf{1.00} & \textbf{1.00} \\
\bottomrule
\end{tabular}
}
\label{tab:task1}
\end{table*}

\begin{figure}
    \centering
    \includegraphics[width=0.90\linewidth]{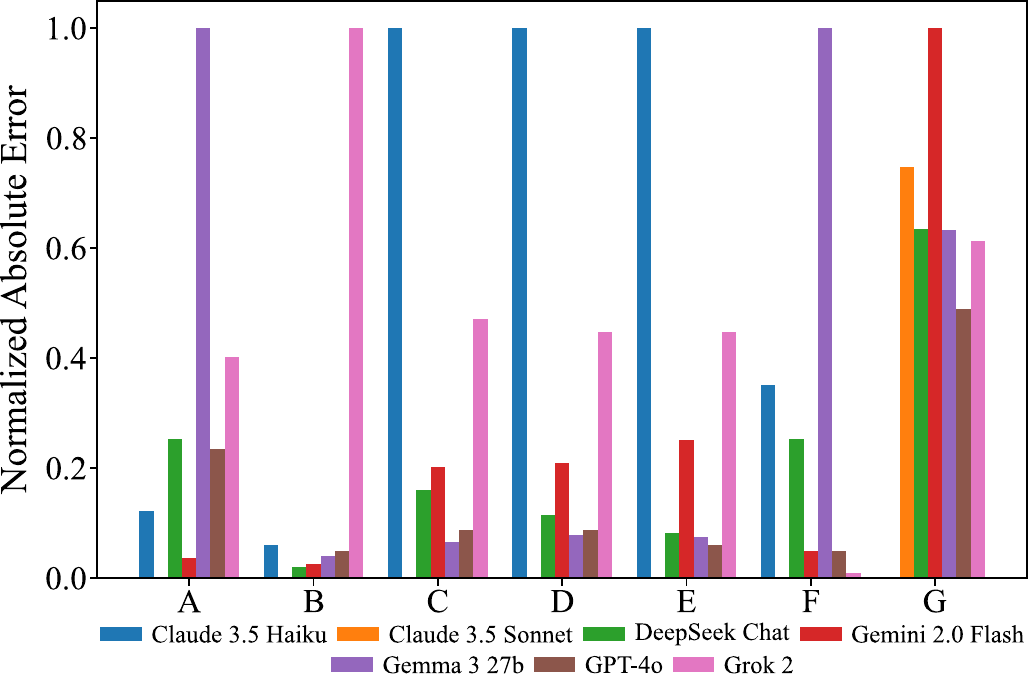}
\caption{Comparison of normalized absolute errors for key structural metrics across models in Task 1. Metrics include: A) Atomic count error, B) Cell volume error, C) Lattice parameter \(a\) error, D) Lattice parameter \(b\) error, E) Lattice parameter \(c\) error, F) Average nearest neighbor distance error, and G) Density error.}
\label{fig:abs_error}
\end{figure}

Table \ref{tab:task1} summarises regression and text metrics; the
normalised absolute-error profile is shown in Figure \ref{fig:abs_error}.  Claude 3.5 Sonnet records the lowest scalar deviation (e.g.,\ $\%\Delta V=44.31$), while Grok 2 inflates cell volumes by over an order of magnitude (\(\%\Delta V=1536.98\)), confirming a unit-scaling failure. BLEU and ROUGE scores cluster around 0.60–0.75, indicating that surface fluency overstates numeric fidelity—a recurrent data-quality pitfall. Image-only ablations raise MAE by 1.7×, underscoring the benefit of multimodal curation.

In Table \ref{tab:task2}, Claude 3.5 Haiku and Gemma 3 27B achieve perfect validity (100 \%) despite moderate atom-count errors (20.34 \% and 24.58 \%, respectively).  Gemini Flash generates valid structures only 55 \% of the time, and DeepSeek Chat records the highest
atom-count error (68.93 \%), suggesting brittle size extrapolation. Qualitative inspection reveals that many invalid Flash samples collapse into thin plates—an artefact that validity and RMSD metrics capture. Cross-chemistry analyses further indicate that FCC Au and Ag are easier to extrapolate than wurtzite ZnO, hinting at symmetry-wise data imbalance.

\textbf{Key observations} (i) Multimodal inputs materially improve lattice inference; (ii) lexical metrics alone cannot guarantee numeric faithfulness, motivating numeracy-aware decoding; (iii) robust generalisation across radii remains challenging, and data-driven augmentation appears a promising remedy.

\begin{table}[ht]
\caption{Performance metrics for Task 2, detailing the percentage of valid generated structures, average RMSD, match rate, and atom count error across various models. Runs are averaged over 10 runs on predicting for R9 of Au material. N/A represents wherever
\(\widehat N_i\neq N_i^{\mathrm{gt}}\) for every test instance. Top performers are highlighted in \textbf{bold}, while runners‑up are \underline{underlined}.}
\label{tab:task2}
\resizebox{\columnwidth}{!}{%
\begin{tabular}{lcccc}
\toprule
Model & Validity (\%) & Avg RMSD & Avg Match Rate (\%) & Avg Atom Count Error (\%) \\
\midrule
Claude 3.5 Haiku & \textbf{100.00} & N/A & N/A & \underline{20.34} \\
Claude 3.5 Sonnet & \underline{90.00} & N/A & N/A & \textbf{19.65} \\
Deepseek Chat & \underline{90.00} & N/A & N/A & 68.93 \\
Gemini 2.0 Flash & 55.00 & N/A & N/A & 64.82 \\
Gemma 3 27b & \textbf{100.00} & N/A & N/A & 24.58 \\
GPT-4o & 85.00 & N/A & N/A & 42.43 \\
\bottomrule
\end{tabular}
}
\end{table}

\section{Limitations}
\label{sec:limitations}

While MCS-Set advances data-centric crystallography, several limitations should be considered when interpreting benchmark results. First, the dataset includes \(\approx 47{,}000\) clusters—sufficient to challenge contemporary multimodal LLMs but still modest by deep learning standards—raising the risk of model memorization rather than generalization. Second, chemical diversity is limited to four inorganic systems (Ag, Au, PbS, ZnO), and all samples are synthetically generated under ideal, noise-free conditions. As a result, the benchmark underrepresents real-world artifacts such as imaging noise, surface reconstruction, and non-stoichiometric defects. Third, evaluation focuses on geometry-aware metrics and numerical fidelity, omitting checks for energetic plausibility or downstream tasks such as DFT relaxation. This gap may permit models to generate geometrically valid yet thermodynamically unstable structures. Finally, the generative task targets extrapolation along the size axis only; generalization across composition, lattice symmetry, or temperature remains outside the current scope and will require future dataset extensions and new task formulations.

\section{Discussion \& Conclusion}
\label{sec:conclusion}

MCS-Set introduces a fully deterministic data-generation pipeline and two benchmark tasks designed to address key data-centric challenges in materials informatics. Task 1 assesses the extent to which multimodal inputs enhance lattice-scale property prediction and textual summary generation. Task 2 probes structural extrapolation under controlled size variations, emphasizing model robustness. Each task includes standardized evaluation metrics and failure-mode slicing tools to support transparent, reproducible benchmarking.

Baseline results across seven large language models reveal modality-specific effects. Incorporating image inputs reduces mean absolute error on geometric scalars by nearly a factor of two, demonstrating that aligned visual cues provide information not recoverable from coordinates alone. In contrast, high BLEU and ROUGE scores often coincide with low numeric FactScores, suggesting that textual fluency does not reliably reflect scientific accuracy. In the generative setting, only two models maintain perfect structural validity, while atom-count error remains close to 20\%, highlighting the ongoing difficulty of extrapolating to larger, less symmetric clusters.

Future directions include numeracy-aware decoding strategies, uncertainty-calibrated objectives, and physics-guided post-relaxation procedures. These extensions aim to close fidelity gaps and improve understanding of how data quality, modality alignment, and evaluation design jointly influence progress in data-centric crystallography. MCS-Set establishes a foundational framework for multimodal, human-in-the-loop data curation, contributing to a more accessible and systematic approach to materials informatics.

\section*{Software and Data}

All data and code used in this work are publicly available. The MCS-Set dataset, along with the benchmark tasks, evaluation metrics, and baseline model implementations, can be accessed at: \url{https://github.com/KurbanIntelligenceLab/MultiCrystalSpectrumSet}

\section*{Impact Statement}

This work introduces an open, audit-ready benchmark designed to support data-centric research in computational crystallography and materials discovery. Expected benefits include accelerated screening of functional materials such as battery and catalyst candidates, reduced dependence on computationally expensive \emph{ab initio} methods, and improved understanding of multimodal learning under domain-specific constraints. Potential risks include the amplification of dataset biases by machine learning models and the unintended generation of unrealistic or chemically unsafe crystal structures.

\bibliography{main}
\bibliographystyle{icml2025}




\end{document}